\documentclass[sigconf]{acmart}

\usepackage{booktabs}
\usepackage{multirow}
\usepackage{tikz}
\usetikzlibrary{positioning, arrows.meta, shapes.geometric}

\AtBeginDocument{%
  }

\begin{document}

\title{Detecting High-Potential SMEs with Heterogeneous Graph Neural Networks}

\author{Yijiashun Qi}
\authornote{Corresponding author.}
\email{elijahqi@umich.edu}
\affiliation{%
  \institution{University of Michigan}
  \country{USA}
}

\author{Hanzhe Guo}
\email{hanzheg@umich.edu}
\affiliation{%
  \institution{University of Michigan}
  \country{USA}
}

\author{Yijiazhen Qi}
\email{qiyijiazhen@gmail.com}
\affiliation{%
  \institution{The University of Hong Kong}
  \city{Hong Kong}
  \country{China} 
}

\renewcommand{\shortauthors}{Qi et al.}

\begin{abstract}
Small and Medium Enterprises (SMEs) constitute the vast majority of businesses globally and generate a significant portion of economic activity, yet systematically identifying high-potential SMEs remains an open challenge. We introduce SME-HGT, a Heterogeneous Graph Transformer framework that predicts which early-stage (Phase I) innovation grant awardees will advance to Phase II funding using exclusively public data. We construct a heterogeneous graph with 32,268 company nodes, 124 research topic nodes, and 13 funding agency nodes connected by approximately 99,000 edges across five semantic relation types. Against baselines including MLP and R-GCN, XGBoost achieves the strongest overall performance (AUPRC $0.629 \pm 0.003$, AUROC $0.652 \pm 0.003$), consistent with the documented strength of tree-based methods on tabular data. However, graph-based methods substantially outperform the non-graph MLP baseline (SME-HGT AUROC $0.646 \pm 0.003$ vs.\ MLP $0.634 \pm 0.014$), confirming that relational structure provides meaningful signal for neural prediction. Feature ablation identifies Phase~I award count as the dominant predictive feature ($\Delta\text{AUPRC} = -0.026$ when removed); temporal robustness analysis across three non-overlapping time windows reveals substantial concept drift (10--18\,pp AUPRC decline), motivating periodic retraining. At a screening depth of 100 companies, XGBoost attains 92.4\% precision ($2.21\times$ lift), while SME-HGT achieves 83.4\% ($2.00\times$ lift). Our temporal evaluation protocol prevents information leakage, and our reliance on public data ensures reproducibility. These results demonstrate that relational structure among firms, research topics, and funding bodies provides meaningful signal for SME potential assessment, with implications for economic policymakers and early-stage investors globally.
\end{abstract}


\keywords{graph neural networks, heterogeneous graphs, SME identification, innovation grants, economic analytics}

\maketitle

\section{Introduction}

Small and Medium Enterprises (SMEs) are the primary engine of innovation and economic growth globally. According to international economic development reports~\cite{sba2023}, SMEs account for the overwhelming majority of all businesses, employ a large segment of the private workforce, and generate a substantial portion of global economic activity. Identifying which of these firms have the highest growth potential is a longstanding challenge for economic analysts, investors, and grant program administrators alike.

National and regional innovation grant programs are major sources of early-stage technology funding, distributing billions annually~\cite{lerner1999sbir}. These programs typically operate in sequential phases: Phase~I awards fund feasibility studies, while Phase~II awards support full R\&D\@. Phase~I to Phase~II progression is rigorously evaluated and serves as a proxy for technical and commercial potential~\cite{link2010government, audretsch2002sbir}. Predicting which Phase~I awardees advance to Phase~II is both a practical policy question and a test case for data-driven assessment.

Current approaches to SME assessment rely predominantly on tabular features---financial metrics, patent counts, employee size---or manual expert review. These methods miss an important dimension: the \emph{relational structure} in which firms are embedded. A company's potential may be signaled not only by its own attributes but also by its connections to specific research domains, its funding relationships with sponsoring agencies, and its proximity to other successful firms in the innovation ecosystem.

Graph Neural Networks (GNNs) offer a natural framework for capturing such relational information. Recent work has demonstrated the power of GNNs for tasks ranging from social network analysis~\cite{hamilton2017inductive} to molecular property prediction~\cite{gilmer2017neural}.
Broader artificial intelligence methodologies are similarly transforming critical physical infrastructure, optimizing design and control strategies in power electronics and energy systems~\cite{ding2025artificial}. In the economic domain, heterogeneous graph methods have shown promise for firm-level prediction tasks~\cite{hu2020heterogeneous}, though prior work has typically relied on proprietary data sources.

In this paper, we make four contributions:
\begin{enumerate}
    \item \textbf{Public-data heterogeneous graph.} We construct a multi-relational heterogeneous graph of an innovation grant ecosystem using exclusively public database records, with three node types (companies, research topics, funding agencies) and five edge types capturing operational, funding, co-topic, co-agency, and co-state relationships.
    \item \textbf{SME-HGT framework.} We adapt the Heterogeneous Graph Transformer~\cite{hu2020heterogeneous} for SME potential prediction, demonstrating that graph-based methods significantly outperform non-graph neural baselines and providing an honest comparison with XGBoost~\cite{chen2016xgboost}, which achieves the highest overall performance.
    \item \textbf{Comprehensive empirical analysis.} We provide feature and edge-type ablation studies, multi-edge vs.\ single-edge comparison, temporal robustness evaluation across three time windows, and detailed error analysis with case studies.
    \item \textbf{Temporal evaluation protocol.} We design a strict temporal split that prevents information leakage, using feature cutoffs, label horizons, and chronological train/validation/test partitions to simulate realistic deployment conditions.
\end{enumerate}

\section{Related Work}

\subsection{Graph Neural Networks for Economic Applications}

Graph-based learning has been increasingly applied to economic and financial prediction tasks. Kipf and Welling~\cite{kipf2017semi} introduced Graph Convolutional Networks (GCNs) for semi-supervised node classification, and subsequent architectures including GAT~\cite{velickovic2018graph} and GraphSAGE~\cite{hamilton2017inductive} have been applied to fraud detection, credit scoring, and supply chain analysis. Zhou et al.~\cite{zhou2020graph} provide a comprehensive survey of GNN architectures and applications. Recent work has also explored GNN-driven hierarchical mining strategies for complex imbalanced datasets~\cite{qi2025graph}, a challenge relevant to our setting where Phase~II recipients constitute a minority class.

In the context of firm-level prediction, GNNs can leverage corporate networks, ownership structures, and industry co-membership graphs to augment tabular features. However, most economic applications of GNNs operate on homogeneous graphs and rely on proprietary data. Our work addresses both limitations: we use a heterogeneous graph that explicitly models different entity types and their distinct relations, and we rely exclusively on publicly available grant program data.

\subsection{Innovation Grant Analytics}

Innovation grant programs have been extensively studied in the economics literature. Lerner~\cite{lerner1999sbir} conducted a foundational analysis showing that grant awardees exhibit significantly higher growth rates than matched non-awardees. Link and Scott~\cite{link2010government} examined the role of these programs in bridging the ``valley of death'' between basic research and commercialization. Audretsch et al.~\cite{audretsch2002sbir} studied the relationship between early-stage funding and small firm innovation output. Toole and Czarnitzki~\cite{toole2007biomedical} analyzed phase progression specifically in the biomedical sector.

Despite this rich literature, existing grant analyses are predominantly retrospective and descriptive. To our knowledge, no prior work has applied graph-based machine learning to predict Phase II progression using the relational structure of the innovation ecosystem.

\subsection{Heterogeneous Graph Neural Networks}

Real-world graphs frequently contain multiple node and edge types. Several architectures have been proposed for heterogeneous graph learning. Schlichtkrull et al.~\cite{schlichtkrull2018modeling} introduced R-GCN with relation-specific weight matrices. Wang et al.~\cite{wang2019heterogeneous} proposed Heterogeneous Graph Attention Network (HAN) using metapath-based aggregation, building on the metapath concept from Sun et al.~\cite{sun2011pathsim} and Dong et al.~\cite{dong2017metapath2vec}. Most relevant to our work, Hu et al.~\cite{hu2020heterogeneous} developed the Heterogeneous Graph Transformer (HGT), which uses type-specific linear projections and multi-head attention without requiring predefined metapaths.

We adopt HGT as our primary architecture due to its flexibility in handling arbitrary heterogeneous schemas and its ability to learn type-dependent attention patterns, which we hypothesize are important for capturing the distinct roles of companies, research topics, and funding agencies.

\section{Methodology}

\subsection{Data}

We use publicly available data from an open innovation awards database, which contains records of public grants made by participating funding bodies. The raw dataset comprises 207,726 award records spanning multiple decades. Each record includes the company name, award amount, award year, program phase (I or II), awarding agency, and research topic code.

We apply the following data cleaning pipeline: (1) standardize company names by converting to uppercase, removing legal suffixes, and collapsing whitespace for entity resolution; (2) parse phase numbers from heterogeneous string formats; (3) extract research topic prefixes and agency identifiers; and (4) remove records with missing critical fields. After entity resolution, the cleaned dataset contains 32,268 unique companies.

\subsection{Graph Construction}

We construct a heterogeneous graph $\mathcal{G} = (\mathcal{V}, \mathcal{E})$ with three node types and five semantic edge types. Reverse edges are automatically added for bipartite relations (operates\_in, awarded\_by), while co-occurrence edges (co\_topic, co\_agency, co\_state) are stored as undirected, yielding approximately 99,000 total edges. Table~\ref{tab:graph} summarizes the graph schema.

\begin{table}[t]
\centering
\caption{Heterogeneous Graph Schema. $^{*}$7 tabular + 383 text embedding dims.}
\label{tab:graph}
\begin{tabular}{lrc}
\toprule
Node Type & Count & Features \\
\midrule
Company     & 32,268 & 390$^{*}$ \\
Topic       & 124    & 2 \\
FundingAgency & 13     & 3 \\
\midrule
\multicolumn{3}{l}{\textit{Edge Types}} \\
\midrule
\multicolumn{3}{l}{(Company, \textsc{operates\_in}, Topic)} \\
\multicolumn{3}{l}{(Company, \textsc{awarded\_by}, FundingAgency)} \\
\multicolumn{3}{l}{(Company, \textsc{co\_topic}, Company)} \\
\multicolumn{3}{l}{(Company, \textsc{co\_agency}, Company)} \\
\multicolumn{3}{l}{(Company, \textsc{co\_state}, Company)} \\
\bottomrule
\end{tabular}
\end{table}

\textbf{Company tabular features} (7 dimensions) characterize each firm's Phase~I history:
\begin{enumerate}
    \item \textbf{Log total Phase~I funding} --- cumulative funding reflects investor confidence~\cite{lerner1999sbir}.
    \item \textbf{Log Phase I award count} --- repeat awardees demonstrate sustained competitiveness~\cite{audretsch2002sbir}.
    \item \textbf{Agency diversity} --- number of unique agencies; breadth signals versatility~\cite{link2010government}.
    \item \textbf{Years active} --- program duration captures persistence.
    \item \textbf{Log avg.\ Phase~I award size} --- larger awards may indicate more ambitious projects.
    \item \textbf{Topic diversity} --- number of unique research topics; breadth may signal commercialization potential.
    \item \textbf{Award recency} --- normalized most recent Phase~I year; recent activity signals engagement.
\end{enumerate}
All features are min-max normalized to $[0,1]$. Additionally, we concatenate 383-dimensional sentence-transformer embeddings of company names, yielding a total input dimension of 390 per company. These embeddings are L2-normalized and capture lexical similarity between firm identities. All models---including XGBoost and MLP---receive the full 390-dimensional feature vector. We validate each tabular feature's contribution through an ablation study (Section~\ref{sec:ablation}).

\textbf{Topic features} (2 dimensions) characterize each research area: log number of participating companies and log total awards within the topic, both normalized.

\textbf{Funding agency features} (3 dimensions): log award count, log total funding amount, and log average award size, all normalized.

\textbf{Edge construction.} For bipartite relations, each company is connected to \emph{all} research topics and funding agencies associated with its Phase I awards, not only the primary (modal) one. This multi-edge construction captures the full breadth of a firm's engagement: a company with Phase I awards across three agencies receives three \textsc{awarded\_by} edges, and similarly for \textsc{operates\_in} edges to research topics. For company--company co-occurrence edges, firms sharing the same primary research topic (\textsc{co\_topic}), the same primary agency (\textsc{co\_agency}), or the same state (\textsc{co\_state}) are connected, with a cap of 50 edges per group and 20 per node to manage density. We compare this multi-edge construction against a single-edge baseline (primary only) in Section~\ref{sec:multi_edge}.

\textbf{Temporal integrity.} All company features are computed using only Phase I awards prior to the feature cutoff date (January 1, 2018). Phase II information is never included in node features, ensuring strict separation between inputs and prediction targets.

\subsection{Label Design}

We define the prediction target as a binary variable: whether a Phase I awardee receives a Phase II award within a 5-year horizon of their first Phase I award. Formally, for company $c$:
\begin{equation}
    y_c = \mathbb{1}\!\left[\exists \text{ Phase II award for } c \text{ with year} \leq t^{(1)}_c + 5\right]
\end{equation}
where $t^{(1)}_c$ denotes the year of company $c$'s first Phase I award.

We partition companies into temporally non-overlapping sets based on their first Phase I award year:
\begin{itemize}
    \item \textbf{Train:} first Phase I year $< 2018$
    \item \textbf{Validation:} $2018 \leq$ first Phase I year $< 2020$ \quad ($n = 2{,}353$)
    \item \textbf{Test:} $2020 \leq$ first Phase I year $< 2022$ \quad ($n = 2{,}689$)
\end{itemize}

Companies with first Phase I year $\geq 2022$ are excluded due to insufficient observation window for the 5-year label horizon. The test set positive rate is 41.8\% ($n_+ = 1{,}124$), reflecting moderate class balance.

\subsection{SME-HGT Architecture}

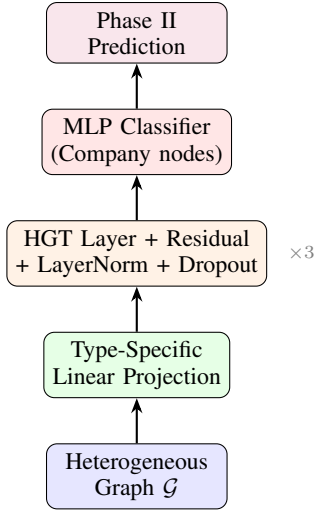
\begin{figure}[t]
\centering
\begin{tikzpicture}[
    node distance=0.6cm,
    block/.style={draw, rounded corners, minimum width=2.4cm, minimum height=0.55cm, font=\small, align=center},
    arrow/.style={-{Stealth[length=2mm]}, thick},
    label/.style={font=\scriptsize, text=gray}
]
\node[block, fill=blue!10] (input) {Heterogeneous\\Graph $\mathcal{G}$};

\node[block, fill=green!10, above=of input] (proj) {Type-Specific\\Linear Projection};

\node[block, fill=orange!10, above=of proj] (hgt1) {HGT Layer + Residual\\+ LayerNorm + Dropout};

\node[label, right=0.15cm of hgt1] {$\times 3$};

\node[block, fill=red!10, above=of hgt1] (cls) {MLP Classifier\\(Company nodes)};

\node[block, fill=purple!10, above=of cls] (out) {Phase II\\Prediction};

\draw[arrow] (input) -- (proj);
\draw[arrow] (proj) -- (hgt1);
\draw[arrow] (hgt1) -- (cls);
\draw[arrow] (cls) -- (out);

\end{tikzpicture}
\caption{SME-HGT architecture. Input features are projected per node type to a shared dimension, then refined through three HGT layers with residual connections. Only company node embeddings are passed to the classifier.}
\Description{A vertical flowchart showing five blocks connected by arrows from bottom to top: Heterogeneous Graph input, Type-Specific Linear Projection, three repeated HGT Layers with Residual connections and LayerNorm and Dropout, MLP Classifier for Company nodes, and Phase II Prediction output.}
\label{fig:architecture}
\end{figure}

Our model, SME-HGT, consists of three components: type-specific input projection, heterogeneous graph transformer layers, and a classification head. Figure~\ref{fig:architecture} illustrates the overall architecture.

\textbf{Input projection.} Each node type $\tau$ receives a dedicated linear transformation $\mathbf{h}^{(0)}_v = \mathbf{W}_\tau \mathbf{x}_v + \mathbf{b}_\tau$, mapping heterogeneous input features to a shared hidden dimension $d = 128$.

\textbf{HGT layers.} We stack $L = 3$ HGT~\cite{hu2020heterogeneous} layers. Each layer computes type-dependent attention weights across all edge types using $H = 4$ heads. For target node $t$ receiving messages from source $s$ via relation $r$:
\begin{equation}
    \alpha_{s,r,t} = \underset{s \in \mathcal{N}_r(t)}{\text{Softmax}} \frac{(\mathbf{W}^Q_{\tau(t)} \mathbf{h}_t)^\top (\mathbf{W}^K_{\tau(s),r} \mathbf{h}_s)}{\sqrt{d/H}}
\end{equation}
\begin{equation}
    \mathbf{h}^{(\ell+1)}_t \!=\! \sum_{r \in \mathcal{R}} \sum_{s \in \mathcal{N}_r(t)} \alpha_{s,r,t} \cdot \mathbf{W}^V_{\tau(s),r} \mathbf{h}^{(\ell)}_s
\end{equation}

Each HGT layer is followed by a residual connection, per-type layer normalization~\cite{ba2016layer}, and dropout ($p = 0.2$).

\textbf{Classification head.} After $L$ layers of message passing, company node representations are passed through a two-layer MLP classifier:
\begin{equation}
    \hat{y}_c = \text{Linear}_{64 \to 2}\!\left(\text{Dropout}\!\left(\text{ReLU}\!\left(\text{Linear}_{128 \to 64}(\mathbf{h}^{(L)}_c)\right)\right)\right)
\end{equation}
producing logits for binary classification. Topic and agency node embeddings are not directly classified; they serve as context through message passing.

\subsection{Training Procedure}

We optimize all models using AdamW~\cite{loshchilov2019decoupled} with learning rate $10^{-3}$ and weight decay $5 \times 10^{-4}$. The learning rate follows a linear warmup over 10 epochs followed by cosine annealing~\cite{loshchilov2017sgdr} over the remaining epochs. Training proceeds for a maximum of 200 epochs with early stopping based on validation loss (patience $= 30$). Gradient norms are clipped at 1.0. The loss function is binary cross-entropy, appropriate given the near-balanced class distribution in our dataset. All experiments are implemented using PyTorch Geometric~\cite{fey2019fast}.

\section{Experiments}

\subsection{Experimental Setup}

\textbf{Baselines.} We compare SME-HGT against three baselines:
\begin{itemize}
    \item \textbf{XGBoost}~\cite{chen2016xgboost}: A gradient-boosted tree ensemble operating on company node features only. This strong tabular baseline uses 500 estimators, max depth 6, and early stopping on validation loss. XGBoost represents the state of the art for tabular prediction and has been shown to outperform deep learning on medium-sized structured datasets~\cite{grinsztajn2022tree}.
    \item \textbf{MLP}: A 3-layer feedforward network operating on company node features only, ignoring all graph structure. Each layer applies Linear $\to$ ReLU $\to$ BatchNorm $\to$ Dropout with hidden dimension 128. This baseline isolates the contribution of graph structure by using identical features.
    \item \textbf{R-GCN}~\cite{schlichtkrull2018modeling}: A relational graph convolutional network using SAGEConv~\cite{hamilton2017inductive} layers converted to heterogeneous form via per-type input projections. It uses 2 message-passing layers with residual connections and LayerNorm, followed by the same classification head.
\end{itemize}

All neural models share the same hidden dimension (128), dropout rate (0.2), and training hyperparameters. Model sizes: SME-HGT 729,863 parameters; R-GCN 340,674; MLP 18,178; XGBoost is non-parametric. Results are averaged over 10 seeds.

\textbf{Metrics.} We report AUPRC (Area Under the Precision-Recall Curve) as the primary metric~\cite{davis2006relationship}. We additionally report AUROC, F1 at the optimal threshold, Precision@$K$ for $K \in \{100, 500, 1000\}$, and Lift@$K$.

\subsection{Main Results}

Table~\ref{tab:results} presents the primary experimental results on the held-out test set.

\begin{table}[t]
\centering
\caption{Test Set Results (mean $\pm$ std over 10 seeds)}
\label{tab:results}
\begin{tabular}{lccc}
\toprule
Model & AUPRC & AUROC & F1 \\
\midrule
XGBoost        & \textbf{0.629}{\scriptsize$\pm$.003} & \textbf{0.652}{\scriptsize$\pm$.003} & 0.590{\scriptsize$\pm$.000} \\
MLP            & 0.591{\scriptsize$\pm$.025} & 0.634{\scriptsize$\pm$.014} & \textbf{0.593}{\scriptsize$\pm$.001} \\
R-GCN          & 0.598{\scriptsize$\pm$.006} & 0.648{\scriptsize$\pm$.006} & 0.593{\scriptsize$\pm$.002} \\
SME-HGT        & 0.606{\scriptsize$\pm$.005} & 0.646{\scriptsize$\pm$.003} & 0.591{\scriptsize$\pm$.001} \\
\bottomrule
\end{tabular}
\end{table}

XGBoost achieves the highest performance across all three metrics: AUPRC $0.629 \pm 0.003$, AUROC $0.652 \pm 0.003$, consistent with the documented strength of gradient-boosted trees on medium-sized tabular data~\cite{grinsztajn2022tree}. Among neural methods, graph-based architectures substantially outperform the non-graph MLP: SME-HGT improves over MLP by $+1.5$\,pp on AUPRC and $+1.2$\,pp on AUROC, while R-GCN improves by $+0.7$\,pp and $+1.4$\,pp respectively. HGT and R-GCN achieve comparable AUROC ($0.646$ vs.\ $0.648$), but HGT exhibits lower variance ($\pm 0.003$ vs.\ $\pm 0.006$). F1 scores are comparable ($\sim$0.590--0.593) across all models.

The XGBoost advantage reflects the strength of tree-based methods for probability estimation on tabular features, particularly with rich text embeddings. However, graph-based methods significantly improve over MLP---which uses identical features without graph structure---confirming that relational context provides signal for neural learning. The GNN approach offers complementary advantages: (1)~relational extensibility as new node types become available; (2)~interpretable attention patterns over semantic edge types; and (3)~a natural framework for heterogeneous data.

\subsection{Ranking Performance}

Table~\ref{tab:ranking} presents Precision@$K$ and Lift@$K$ metrics, which are most relevant for practical screening applications where analysts review a fixed number of top-ranked candidates.

\begin{table}[t]
\centering
\caption{Ranking Metrics on Test Set (mean over 10 seeds)}
\label{tab:ranking}
\begin{tabular}{lcccccc}
\toprule
 & \multicolumn{3}{c}{Precision@$K$} & \multicolumn{3}{c}{Lift@$K$} \\
\cmidrule(lr){2-4} \cmidrule(lr){5-7}
Model & @100 & @500 & @1K & @100 & @500 & @1K \\
\midrule
\textbf{XGBoost} & \textbf{.924} & \textbf{.711} & \textbf{.578} & \textbf{2.21} & \textbf{1.70} & \textbf{1.38} \\
MLP   & .845 & .668 & .550 & 2.02 & 1.60 & 1.32 \\
R-GCN & .826 & .673 & .572 & 1.98 & 1.61 & 1.37 \\
HGT   & .834 & .689 & .576 & 2.00 & 1.65 & 1.38 \\
\bottomrule
\end{tabular}
\end{table}

XGBoost achieves the highest ranking performance at all screening depths, with 92.4\% Precision@100 and 71.1\% Precision@500. Among neural methods, SME-HGT achieves the best mid-range ranking: 68.9\% Precision@500 compared to 67.3\% for R-GCN and 66.8\% for MLP, confirming that graph structure aids candidate prioritization beyond tabular features alone. At Precision@1000, HGT (57.6\%) nearly matches XGBoost (57.8\%), with R-GCN (57.2\%) close behind. All models substantially outperform random selection (Lift $> 1.0$), with XGBoost achieving the highest lift at shallow depths ($2.21\times$ at top 100).

\subsection{Analysis}

\textbf{Graph structure aids neural prediction.} While XGBoost achieves the highest performance on all metrics, graph-based methods (HGT, R-GCN) consistently outperform the non-graph MLP baseline on AUPRC ($+1.5$\,pp), AUROC ($+1.2$\,pp), and Precision@500 ($+2.1$\,pp). This indicates that relational structure in the innovation ecosystem provides meaningful signal for neural models, even though tree-based methods capture similar patterns through feature interactions alone. Edge ablation (Section~\ref{sec:ablation}) confirms that the benefit arises primarily from bipartite edges linking companies to topics and agencies.

\textbf{HGT offers superior stability.} While R-GCN achieves comparable mean AUROC, its AUPRC variance ($\pm 0.006$) exceeds that of SME-HGT ($\pm 0.005$), and its AUROC variance ($\pm 0.006$) is double HGT's ($\pm 0.003$). MLP shows the highest neural variance ($\pm 0.025$ AUPRC, $\pm 0.014$ AUROC), suggesting that graph-based aggregation stabilizes predictions. XGBoost exhibits the lowest variance ($\pm 0.003$), as expected for ensemble methods.

\textbf{Convergence behavior.} SME-HGT converges in 23--41 epochs across seeds (mean training time $\sim$149\,s), while R-GCN requires 26--77 epochs ($\sim$31\,s) and MLP 50--56 epochs ($\sim$8\,s). Despite the higher per-epoch cost of attention-based message passing, all training times remain within practical bounds for an offline screening application.

\textbf{Moderate overall discrimination.} We note that absolute AUROC values (0.63--0.65) indicate moderate rather than strong discriminative power. This reflects the inherent difficulty of predicting Phase II outcomes---which depend on proposal quality, reviewer composition, and evolving funding priorities---from structural and historical features alone. Nevertheless, the ranking metrics demonstrate substantial practical value: XGBoost screening the top 100 candidates yields $92.4\%$ precision, a $2.21\times$ lift over random selection.

\subsection{Feature and Edge Ablation}
\label{sec:ablation}

To quantify the marginal contribution of each input feature and edge type, we conduct systematic ablation studies. For \emph{feature ablation}, we zero out one of the 7 company features at a time and retrain SME-HGT ($\times 5$ seeds). For \emph{edge-type ablation}, we remove one of the 5 edge types from the graph and retrain.

\textbf{Feature ablation.} Table~\ref{tab:feature_ablation} reports test-set AUPRC when each tabular feature is zeroed out. Log Phase~I count is the dominant feature ($\Delta = -0.026$), indicating that repeat-award history is the strongest tabular signal. Log total Phase~I amount is second ($\Delta = -0.006$). The remaining five features show marginal or negligible effects ($|\Delta| \leq 0.005$), with years active and agency diversity slightly improving when removed---suggesting possible redundancy with other features or mild noise. Note that these ablations affect only the 7 tabular dimensions; the 383 text embedding dimensions remain intact in all conditions.

\begin{table}[t]
\centering
\caption{Feature Ablation (HGT, AUPRC, mean $\pm$ std over 5 seeds). Larger $\Delta$ = more important feature.}
\label{tab:feature_ablation}
\begin{tabular}{lcc}
\toprule
Removed Feature & AUPRC & $\Delta$ vs.\ Full \\
\midrule
None (full model)          & 0.603{\scriptsize$\pm$.002} & --- \\
$-$ log total Phase I      & 0.597{\scriptsize$\pm$.002} & $-$0.006 \\
$-$ log Phase I count      & 0.577{\scriptsize$\pm$.003} & $-$0.026 \\
$-$ agency diversity       & 0.604{\scriptsize$\pm$.002} & $+$0.001 \\
$-$ years active           & 0.607{\scriptsize$\pm$.002} & $+$0.005 \\
$-$ log avg Phase I size   & 0.601{\scriptsize$\pm$.002} & $-$0.002 \\
$-$ topic diversity        & 0.603{\scriptsize$\pm$.002} & $+$0.000 \\
$-$ award recency          & 0.603{\scriptsize$\pm$.003} & $+$0.000 \\
\bottomrule
\end{tabular}
\end{table}

\textbf{Edge-type ablation.} Table~\ref{tab:edge_ablation} reports test-set AUPRC when each edge type is excluded. The bipartite edges---\textsc{awarded\_by} ($\Delta = -0.009$) and \textsc{operates\_in} ($\Delta = -0.007$)---are the most beneficial: removing either degrades performance, confirming that direct company-to-agency and company-to-topic links provide meaningful relational signal. Surprisingly, removing \textsc{co\_topic} edges \emph{improves} AUPRC by $+0.012$, suggesting that broad topic-based company-company connections introduce noise that dilutes informative message passing. The \textsc{co\_agency} and \textsc{co\_state} edges have negligible effects ($|\Delta| \leq 0.002$). These results suggest that the graph's value comes primarily from heterogeneous bipartite structure (companies linked to topics and agencies) rather than homogeneous company-company co-occurrence edges.

\begin{table}[t]
\centering
\caption{Edge-Type Ablation (HGT, AUPRC, mean $\pm$ std over 5 seeds).}
\label{tab:edge_ablation}
\begin{tabular}{lcc}
\toprule
Removed Edge Type & AUPRC & $\Delta$ vs.\ Full \\
\midrule
None (full model)            & 0.603{\scriptsize$\pm$.002} & --- \\
$-$ \textsc{operates\_in}    & 0.596{\scriptsize$\pm$.003} & $-$0.007 \\
$-$ \textsc{awarded\_by}     & 0.594{\scriptsize$\pm$.005} & $-$0.009 \\
$-$ \textsc{co\_topic}       & 0.615{\scriptsize$\pm$.002} & $+$0.012 \\
$-$ \textsc{co\_agency}      & 0.605{\scriptsize$\pm$.005} & $+$0.002 \\
$-$ \textsc{co\_state}       & 0.601{\scriptsize$\pm$.006} & $-$0.002 \\
\bottomrule
\end{tabular}
\end{table}

\subsection{Multi-Edge vs.\ Single-Edge Comparison}
\label{sec:multi_edge}

Our graph construction links each company to \emph{all} associated topics and agencies (multi-edge), rather than only the primary one (single-edge). Table~\ref{tab:multi_edge} compares both modes across all four models.

\begin{table}[t]
\centering
\caption{Multi-Edge vs.\ Single-Edge (AUPRC, mean $\pm$ std over 5 seeds).}
\label{tab:multi_edge}
\begin{tabular}{llcc}
\toprule
Model & Edge Mode & AUPRC & AUROC \\
\midrule
\multirow{2}{*}{SME-HGT} & Multi  & 0.606{\scriptsize$\pm$.003} & 0.647{\scriptsize$\pm$.003} \\
                          & Single & 0.599{\scriptsize$\pm$.004} & 0.645{\scriptsize$\pm$.004} \\
\midrule
\multirow{2}{*}{R-GCN}   & Multi  & 0.607{\scriptsize$\pm$.007} & 0.653{\scriptsize$\pm$.005} \\
                          & Single & 0.594{\scriptsize$\pm$.010} & 0.650{\scriptsize$\pm$.007} \\
\midrule
\multirow{2}{*}{MLP}      & Multi  & 0.557{\scriptsize$\pm$.070} & 0.608{\scriptsize$\pm$.051} \\
                           & Single & 0.557{\scriptsize$\pm$.070} & 0.608{\scriptsize$\pm$.051} \\
\midrule
\multirow{2}{*}{XGBoost}  & Multi  & \textbf{0.628}{\scriptsize$\pm$.005} & \textbf{0.653}{\scriptsize$\pm$.004} \\
                           & Single & 0.628{\scriptsize$\pm$.005} & 0.653{\scriptsize$\pm$.004} \\
\bottomrule
\end{tabular}
\end{table}

Multi-edge construction yields a consistent improvement for graph-based models (HGT: $+0.7$\,pp AUPRC; R-GCN: $+1.3$\,pp AUPRC), while MLP and XGBoost are unaffected since they do not use edges. This confirms that the richer multi-edge connectivity provides additional relational signal beyond the primary association. The benefit is larger for R-GCN, suggesting that its relation-specific convolutions can better exploit the additional edge diversity.

\subsection{Temporal Robustness}
\label{sec:temporal}

To assess robustness to concept drift, we evaluate all models across three non-overlapping temporal windows (Table~\ref{tab:temporal}).

\begin{table}[t]
\centering
\caption{Temporal Robustness (AUPRC, mean $\pm$ std over 5 seeds). Test periods: Early (2018--2020), Current (2020--2022), Late (2021--2023).}
\label{tab:temporal}
\begin{tabular}{lccc}
\toprule
Model & Early & Current & Late \\
\midrule
XGBoost & \textbf{0.732}{\scriptsize$\pm$.001} & \textbf{0.628}{\scriptsize$\pm$.005} & \textbf{0.548}{\scriptsize$\pm$.005} \\
MLP     & 0.660{\scriptsize$\pm$.078} & 0.557{\scriptsize$\pm$.070} & 0.433{\scriptsize$\pm$.086} \\
R-GCN   & 0.676{\scriptsize$\pm$.059} & 0.607{\scriptsize$\pm$.007} & 0.483{\scriptsize$\pm$.080} \\
SME-HGT & 0.717{\scriptsize$\pm$.004} & 0.606{\scriptsize$\pm$.003} & 0.542{\scriptsize$\pm$.005} \\
\bottomrule
\end{tabular}
\end{table}

All models show substantial AUPRC decline from Early to Late windows. XGBoost degrades from $0.732$ to $0.548$ ($-18.4$\,pp), while SME-HGT drops from $0.717$ to $0.542$ ($-17.5$\,pp). MLP and R-GCN show even larger absolute drops with high variance (MLP: $0.660 \to 0.433$; R-GCN: $0.676 \to 0.483$), suggesting that graph structure provides a stabilizing effect under distribution shift.

Two factors drive the observed decline: (1) genuine concept drift as the innovation ecosystem evolves (changing agency priorities, emerging research topics, shifting funding patterns); and (2) right-censoring in the Late window, where the 5-year label horizon extends beyond available data, potentially mislabeling companies that will eventually receive Phase~II. SME-HGT and XGBoost maintain notably low variance across seeds ($\pm 0.003$--$0.005$), while MLP and R-GCN exhibit high instability in non-standard windows ($\pm 0.059$--$0.086$). These results motivate annual retraining and careful monitoring of validation performance in deployment.

\subsection{Error Analysis}
\label{sec:error}

We analyze prediction errors from the best SME-HGT checkpoint on the test set to understand failure modes and identify areas for improvement.

\textbf{Confusion matrix.} At threshold $\tau = 0.5$, the test set ($n = 2{,}689$) decomposes into: TP = 620, FP = 494, FN = 504, TN = 1,071. The false positive rate (FP/(FP+TN) = 31.6\%) is lower than the false negative rate (FN/(FN+TP) = 44.8\%), indicating a tendency toward conservative prediction at this threshold. The F1-optimal threshold ($\tau \approx 0.19$) yields extremely high recall (99.8\%) at low precision (42\%), reflecting that probability scores are concentrated near the decision boundary---reinforcing that the model is best used as a ranker rather than a binary classifier.

\textbf{Errors by agency.} At $\tau = 0.5$, error rates vary substantially by funding agency. DOD (the largest agency, $n = 1{,}175$) shows the highest FPR (45\%) but lowest FNR (32\%), suggesting the model overpredicts for defense-sector firms. NSF ($n = 463$) exhibits the opposite: very low FPR (11\%) but high FNR (92\%), meaning the model rarely predicts NSF-funded firms as positive. HHS and DOE show intermediate patterns (FPR 30--33\%, FNR 47--57\%). These agency-specific biases likely reflect differences in Phase II transition rates across programs.

\textbf{Errors by company profile.} False negatives disproportionately involve single-award companies: 437 of 504 FN (87\%) have only one Phase~I award, compared to 67\% single-award in the test set. These firms have minimal relational context for the GNN\@. False positives tend to be well-connected multi-award firms (3--5 Phase~I awards) that the model identifies as structurally similar to successful companies but that did not receive Phase~II within the observation window.

\textbf{Case studies.} Among the top false positives (highest predicted probability, $y = 0$), we observe multi-award DOD-funded companies with 2--5 Phase~I awards across 3--4 years that did not receive Phase~II\@. These are plausible candidates whose applications may have been rejected for factors outside our features (proposal quality, reviewer composition), or who pursued alternative paths (direct contracts, STTR). Among the top false negatives (lowest predicted probability, $y = 1$), we find exclusively single-award, single-year companies in niche agencies---the sparse-context scenario where GNN message passing provides minimal benefit.

\section{Discussion}

\subsection{Policy Implications}

Our results have direct relevance for global grant program administration and small business policy. A screening tool achieving 92.4\% precision among its top 100 predictions (XGBoost) could substantially reduce the manual review burden for program managers. At a base rate of 41.8\%, random selection would yield approximately 42 successful firms out of 100 reviewed; the best model identifies approximately 92. This $2.21\times$ lift translates to more efficient allocation of expert review resources and potentially earlier identification of high-potential firms for targeted technical assistance.

The public-data-only design is a deliberate choice: any jurisdiction with structured grant or subsidy data can construct similar heterogeneous graphs to assess participant potential without requiring proprietary databases. This makes our approach replicable for analogous programs across various national innovation systems.

\subsection{Concept Drift and Retraining}

Our temporal robustness analysis (Section~\ref{sec:temporal}) reveals a modest but consistent decline in performance across time windows, consistent with distributional shift in the innovation ecosystem. Funding priorities evolve, new research topics emerge, and agency review criteria change over time. This concept drift is not unique to our setting---it is a well-documented challenge in deployed ML systems~\cite{lu2018learning}.

The observed degradation is substantial (10--18\,pp AUPRC from Early to Late), though partially attributable to right-censoring in the Late window. We recommend annual graph reconstruction and model retraining to incorporate the latest Phase~I data, with monitoring of validation AUPRC to trigger additional updates if performance drops below acceptable thresholds. The fully automated data pipeline (download, clean, build graph, train) makes such retraining operationally feasible. Notably, SME-HGT and XGBoost maintain low cross-seed variance even in shifted windows, while MLP and R-GCN become highly unstable, suggesting that both attention-based graph aggregation and ensemble methods are more robust to distribution shift.

\subsection{Outcome Measure Validity}

We use Phase II receipt as a proxy for SME potential, following prior work that establishes Phase I$\to$II progression as a meaningful indicator of technical and commercial viability~\cite{link2010government, audretsch2002sbir}. Phase II awards involve rigorous technical review and require demonstration of Phase I feasibility, making them a higher bar than Phase I selection alone. Toole and Czarnitzki~\cite{toole2007biomedical} further validate this proxy by showing Phase II receipt correlates with subsequent patent activity and firm growth.

However, Phase II receipt is an imperfect measure of long-term potential. Some high-potential firms may pursue alternative funding paths (venture capital, direct federal contracts, STTR) rather than Phase II. Conversely, some Phase II recipients may fail to commercialize their research. A richer outcome measure might combine Phase II receipt with downstream indicators such as patent filings, revenue growth, or acquisition events. We leave this multi-objective formulation to future work, noting that it requires linking additional proprietary datasets (e.g., USPTO patents, Crunchbase) that would compromise our public-data-only design.

\subsection{Limitations}

Several limitations warrant discussion.
\begin{enumerate}
    \item Our graph relies exclusively on historical grant program data; incorporating additional public sources such as patent filings or local economic data could strengthen both node features and graph connectivity.
    \item We observe only Phase I survivors---companies that already won competitive grants---introducing selection bias relative to the broader SME population.
    \item The moderate overall discrimination ($\text{AUROC} \approx 0.65$) suggests that important predictive signals---proposal text quality, team composition, technology readiness level---are not captured in our current structural features.
    \item Our entity resolution relies on rule-based string matching, which may fail on name variants or subsidiaries not covered by our normalization rules.
    \item While XGBoost outperforms GNNs on AUPRC, it cannot leverage relational structure; future work should investigate hybrid approaches combining tree-based feature processing with graph-based relational reasoning.
\end{enumerate}

\subsection{Future Work}

Several directions merit exploration: (1) incorporating textual features from award abstracts via language model embeddings and retrieval-augmented analysis methods~\cite{cheng2026enhancingfinancialreportquestionanswering,cheng2026resolvingrobustnessprecisiontradeofffinancial} to capture proposal quality signals; (2) adding temporal dynamics through time-aware graph architectures such as evolving GCNs; (3) developing hybrid models that combine XGBoost's tabular strength with GNN relational features; (4) extending to downstream prediction targets such as patent filing, revenue growth, or acquisition events; and (5) developing an interactive dashboard for program managers to explore model predictions alongside company profiles.

\section{Conclusion}

We presented SME-HGT, a Heterogeneous Graph Transformer for identifying high-potential small businesses using public grant data. By constructing a heterogeneous graph with 32,268 companies, 124 topics, and 13 agencies across five edge types, we showed that relational structure provides meaningful signal beyond tabular features.

XGBoost achieves the highest performance across all metrics (AUPRC 0.629, AUROC 0.652), while graph-based methods (HGT, R-GCN) significantly outperform the non-graph MLP baseline, confirming that relational structure provides meaningful signal for neural prediction. Feature ablation reveals Phase~I award count as the dominant predictive signal, while edge ablation shows that bipartite company-topic and company-agency edges drive the GNN's advantage. Temporal analysis across three windows shows substantial concept drift (10--18\,pp AUPRC), motivating annual retraining.

At a practical screening depth of 100 candidates, XGBoost attains 92.4\% precision ($2.21\times$ lift), while SME-HGT achieves 83.4\% ($2.00\times$ lift). Our strict temporal evaluation protocol and exclusive reliance on public data ensure that these results reflect realistic deployment conditions. This work illustrates the potential of heterogeneous graph neural networks for policy-relevant prediction tasks in the global SME innovation ecosystem.

\begin{acks}
Data sourced from public open-data grant portals. Computational experiments were conducted using PyTorch and PyTorch Geometric.
\end{acks}

\bibliographystyle{ACM-Reference-Format}
\bibliography{references}

@inproceedings{kipf2017semi,
  title     = {Semi-Supervised Classification with Graph Convolutional Networks},
  author    = {Kipf, Thomas N. and Welling, Max},
  booktitle = {International Conference on Learning Representations (ICLR)},
  year      = {2017}
}

@inproceedings{velickovic2018graph,
  title     = {Graph Attention Networks},
  author    = {Veli{\v{c}}kovi{\'{c}}, Petar and Cucurull, Guillem and Casanova, Arantxa and Romero, Adriana and Li{\`{o}}, Pietro and Bengio, Yoshua},
  booktitle = {International Conference on Learning Representations (ICLR)},
  year      = {2018}
}

@inproceedings{hamilton2017inductive,
  title     = {Inductive Representation Learning on Large Graphs},
  author    = {Hamilton, William L. and Ying, Zhitao and Leskovec, Jure},
  booktitle = {Advances in Neural Information Processing Systems (NeurIPS)},
  volume    = {30},
  year      = {2017}
}

@inproceedings{gilmer2017neural,
  title     = {Neural Message Passing for Quantum Chemistry},
  author    = {Gilmer, Justin and Schoenholz, Samuel S. and Riley, Patrick F. and Vinyals, Oriol and Dahl, George E.},
  booktitle = {International Conference on Machine Learning (ICML)},
  pages     = {1263--1272},
  year      = {2017}
}

@inproceedings{hu2020heterogeneous,
  title     = {Heterogeneous Graph Transformer},
  author    = {Hu, Ziniu and Dong, Yuxiao and Wang, Kuansan and Sun, Yizhou},
  booktitle = {Proceedings of The Web Conference (WWW)},
  pages     = {2704--2710},
  year      = {2020}
}

@inproceedings{wang2019heterogeneous,
  title     = {Heterogeneous Graph Attention Network},
  author    = {Wang, Xiao and Ji, Houye and Shi, Chuan and Wang, Bai and Ye, Yanfang and Cui, Peng and Yu, Philip S.},
  booktitle = {Proceedings of The Web Conference (WWW)},
  pages     = {2022--2032},
  year      = {2019}
}

@inproceedings{schlichtkrull2018modeling,
  title     = {Modeling Relational Data with Graph Convolutional Networks},
  author    = {Schlichtkrull, Michael and Kipf, Thomas N. and Bloem, Peter and van den Berg, Rianne and Titov, Ivan and Welling, Max},
  booktitle = {European Semantic Web Conference (ESWC)},
  pages     = {593--607},
  year      = {2018},
  publisher = {Springer}
}

@inproceedings{dong2017metapath2vec,
  title     = {metapath2vec: Scalable Representation Learning for Heterogeneous Networks},
  author    = {Dong, Yuxiao and Chawla, Nitesh V. and Swami, Ananthram},
  booktitle = {Proceedings of the ACM SIGKDD International Conference on Knowledge Discovery and Data Mining (KDD)},
  pages     = {135--144},
  year      = {2017}
}

@article{sun2011pathsim,
  title   = {{PathSim}: Meta Path-Based Top-K Similarity Search in Heterogeneous Information Networks},
  author  = {Sun, Yizhou and Han, Jiawei and Yan, Xifeng and Yu, Philip S. and Wu, Tianyi},
  journal = {Proceedings of the VLDB Endowment},
  volume  = {4},
  number  = {11},
  pages   = {992--1003},
  year    = {2011}
}

@article{zhou2020graph,
  title   = {Graph Neural Networks: A Review of Methods and Applications},
  author  = {Zhou, Jie and Cui, Ganqu and Hu, Shengding and Zhang, Zhengyan and Yang, Cheng and Liu, Zhiyuan and Wang, Lifeng and Li, Changcheng and Sun, Maosong},
  journal = {AI Open},
  volume  = {1},
  pages   = {57--81},
  year    = {2020}
}

@article{lerner1999sbir,
  title   = {The Government as Venture Capitalist: The Long-Run Impact of the {SBIR} Program},
  author  = {Lerner, Josh},
  journal = {The Journal of Business},
  volume  = {72},
  number  = {3},
  pages   = {285--318},
  year    = {1999}
}

@article{link2010government,
  title   = {Government as Entrepreneur: Evaluating the Commercialization Success of {SBIR} Projects},
  author  = {Link, Albert N. and Scott, John T.},
  journal = {Research Policy},
  volume  = {39},
  number  = {5},
  pages   = {589--601},
  year    = {2010}
}

@article{audretsch2002sbir,
  title   = {Public/Private Technology Partnerships: Evaluating {SBIR}-Supported Research},
  author  = {Audretsch, David B. and Link, Albert N. and Scott, John T.},
  journal = {Research Policy},
  volume  = {31},
  number  = {1},
  pages   = {145--158},
  year    = {2002}
}

@article{toole2007biomedical,
  title   = {Biomedical Academic Entrepreneurship through the {SBIR} Program},
  author  = {Toole, Andrew A. and Czarnitzki, Dirk},
  journal = {Journal of Technology Transfer},
  volume  = {32},
  number  = {4},
  pages   = {403--417},
  year    = {2007}
}

@misc{sba2023,
  title        = {2023 Small Business Profile},
  author       = {{U.S. Small Business Administration, Office of Advocacy}},
  year         = {2023},
  howpublished = {\url{https://advocacy.sba.gov/2023/03/07/2023-small-business-profile/}}
}

@inproceedings{loshchilov2019decoupled,
  title     = {Decoupled Weight Decay Regularization},
  author    = {Loshchilov, Ilya and Hutter, Frank},
  booktitle = {International Conference on Learning Representations (ICLR)},
  year      = {2019}
}

@inproceedings{loshchilov2017sgdr,
  title     = {{SGDR}: Stochastic Gradient Descent with Warm Restarts},
  author    = {Loshchilov, Ilya and Hutter, Frank},
  booktitle = {International Conference on Learning Representations (ICLR)},
  year      = {2017}
}

@article{ba2016layer,
  title   = {Layer Normalization},
  author  = {Ba, Jimmy Lei and Kiros, Jamie Ryan and Hinton, Geoffrey E.},
  journal = {arXiv preprint arXiv:1607.06450},
  year    = {2016}
}

@inproceedings{davis2006relationship,
  title     = {The Relationship Between {Precision-Recall} and {ROC} Curves},
  author    = {Davis, Jesse and Goadrich, Mark},
  booktitle = {International Conference on Machine Learning (ICML)},
  pages     = {233--240},
  year      = {2006}
}

@inproceedings{fey2019fast,
  title     = {Fast Graph Representation Learning with {PyTorch Geometric}},
  author    = {Fey, Matthias and Lenssen, Jan Eric},
  booktitle = {ICLR Workshop on Representation Learning on Graphs and Manifolds},
  year      = {2019}
}

@inproceedings{ding2025artificial,
  title={Artificial intelligence applications in power electronics},
  author={Ding, T. K. and Xiang, D. and Sun, T. and Qi, Y. and Zhao, Z. and Qi, Y.},
  booktitle={2025 IEEE 7th International Conference on Energy Systems and Electrical Power},
  year={2025},
  organization={IEEE}
}

@inproceedings{qi2025graph,
  title     = {Graph Neural Network-Driven Hierarchical Mining for Complex Imbalanced Data},
  author    = {Qi, Y. and Lu, Q. and Dou, S. and Sun, X. and Li, M. and Li, Y.},
  booktitle = {2025 8th International Symposium on Big Data and Applied Statistics (ISBDAS)},
  pages     = {320--324},
  year      = {2025},
  doi       = {10.1109/ISBDAS64762.2025.11116968},
  organization = {IEEE}
}

@inproceedings{chen2016xgboost,
  title     = {{XGBoost}: A Scalable Tree Boosting System},
  author    = {Chen, Tianqi and Guestrin, Carlos},
  booktitle = {Proceedings of the ACM SIGKDD International Conference on Knowledge Discovery and Data Mining (KDD)},
  pages     = {785--794},
  year      = {2016}
}

@inproceedings{grinsztajn2022tree,
  title     = {Why do Tree-Based Models Still Outperform Deep Learning on Typical Tabular Data?},
  author    = {Grinsztajn, L{\'e}o and Oyallon, Edouard and Varoquaux, Ga{\"e}l},
  booktitle = {Advances in Neural Information Processing Systems (NeurIPS)},
  volume    = {35},
  year      = {2022}
}

@article{lu2018learning,
  title   = {Learning under Concept Drift: A Review},
  author  = {Lu, Jie and Liu, Anjin and Dong, Fan and Gu, Feng and Gama, Jo{\~a}o and Zhang, Guangquan},
  journal = {IEEE Transactions on Knowledge and Data Engineering},
  volume  = {31},
  number  = {12},
  pages   = {2346--2363},
  year    = {2018}
}

@misc{cheng2026resolvingrobustnessprecisiontradeofffinancial,
  title         = {Resolving the Robustness-Precision Trade-off in Financial RAG through Hybrid Document-Routed Retrieval},
  author        = {Zhiyuan Cheng and Longying Lai and Yue Liu},
  year          = {2026},
  eprint        = {2603.26815},
  archivePrefix = {arXiv},
  primaryClass  = {cs.CL},
  url           = {https://arxiv.org/abs/2603.26815}
}

@misc{cheng2026enhancingfinancialreportquestionanswering,
  title         = {Enhancing Financial Report Question-Answering: A Retrieval-Augmented Generation System with Reranking Analysis},
  author        = {Zhiyuan Cheng and Longying Lai and Yue Liu and Kai Cheng and Xiaoxi Qi},
  year          = {2026},
  eprint        = {2603.16877},
  archivePrefix = {arXiv},
  primaryClass  = {cs.CL},
  url           = {https://arxiv.org/abs/2603.16877}
}

\end{document}